\def\year{2020}\relax
\documentclass[letterpaper]{article} 
\usepackage{aaai20}  
\usepackage{times}  
\usepackage{helvet} 
\usepackage{courier}  
\usepackage[hyphens]{url}  
\usepackage{graphicx} 
\usepackage{amsmath}
\usepackage{graphicx}  
\newcommand \newcite[1]{{ \citeauthor{#1}~\shortcite{#1} }}
\newcommand{\tabincell}[2]{
\begin{tabular}{@{}#1@{}}#2\end{tabular}
} 
\usepackage{stmaryrd}
\usepackage{paralist}
\usepackage{subfig}
\usepackage{dashrule}
\usepackage{latexsym}
\usepackage{booktabs}
\usepackage{tabularx}
\usepackage{multirow}
\usepackage{verbatim}
\usepackage{xcolor}
\title{Acquiring Knowledge from Pre-trained Model to Neural Machine Translation}
\author{
\Large \textbf{Rongxiang Weng\textsuperscript{\rm 1,2}\thanks{Code is available at: https://github.com/wengrx/APT-NMT}, Heng Yu\textsuperscript{\rm 2}, Shujian Huang\textsuperscript{\rm 1},}
\\ \Large \textbf{Shanbo Cheng\textsuperscript{\rm 2}, Weihua Luo\textsuperscript{\rm 2}}\\ 
\textsuperscript{\rm 1}National Key Laboratory for Novel Software Technology, Nanjing University, Nanjing, China\\ 
\textsuperscript{\rm 2}Machine Intelligence Technology Lab, Alibaba Group, Hangzhou, China \\ 
\{wengrx,yuheng.yh\}@alibaba-inc.com, huangsj@nju.edu.cn,
\\
\{weihua.luowh\}@alibaba-inc.com 
}
\begin{document}

\maketitle

\begin{abstract}
Pre-training and fine-tuning have achieved great success in natural language process field. 
The standard paradigm of exploiting them includes two steps: first, pre-training a model, e.g. BERT, with a large scale unlabeled monolingual data. Then, fine-tuning the pre-trained model with labeled data from downstream tasks. 
However, in neural machine translation (NMT), we address the problem that the training objective of the bilingual task is far different from the monolingual pre-trained model. This gap leads that only using fine-tuning in NMT can not fully utilize prior language knowledge. 
In this paper, we propose an \textsc{Apt} framework for acquiring knowledge from pre-trained model to NMT. The proposed approach includes two modules: 1). a dynamic fusion mechanism to fuse task-specific features adapted from general knowledge into NMT network, 2). a knowledge distillation paradigm to learn language knowledge continuously during the NMT training process. The proposed approach could integrate suitable knowledge from pre-trained models to improve the NMT.
Experimental results on WMT English to German, German to English and Chinese to English machine translation tasks show that our model outperforms strong baselines and the fine-tuning counterparts.
\end{abstract}
    
\section{Introduction}
Neural machine translation (NMT) based on the \textit{encoder-decoder framework}~\cite{sutskever2014sequence,Cho2014Learning,Bahdanau2015Neural,Luong2015Effective} has obtained state-of-the-art performance on many language pairs~\cite{deng2018alibaba}.
Various advanced neural architectures have been explored for NMT under this framework, such as recurrent neural network (RNN)~\cite[RNNSearch]{Bahdanau2015Neural,Luong2015Effective}, convolutional neural network (CNN)~\cite[Conv-S2S]{gehring2016convolutional} and self-attention network~\cite[Transformer]{vaswani2017attention}.

Currently, most NMT systems only utilize the sentence-aligned parallel corpus for model training.
Monolingual data, which is larger and easier to collect, is not fully utilized limiting the capacity of NMT models.
Previously, several successful attempts have been made to improve NMT by incorporating monolingual data~\cite{gulcehre2015using,sennrich2016improving,Zhang2016Exploiting,poncelas2018investigating}.
However, these studies only focus on the usage of word-level information, e.g. extracting information from word embedding. 
The rich contextual information from large scale monolingual data does not be fully utilized. 
Meanwhile, fine-tuning the parameters from unsupervised pre-trained models, like GPT \cite{radford2018improving} or BERT \cite{devlin2018bert}, 
in which downstream tasks could exploit the contextual knowledge from large scale monolingual data, has gained tremendous success in a variety of natural language process tasks.
Thus, the upcoming question is \textit{whether the contextual knowledge from pre-trained models are useful in NMT}.

Due to the limited amount of high-quality parallel data, NMT as a complex text generation task, can not generate appropriate representation. The contextual knowledge from pre-trained models could naturally be a good complement for NMT. 
Nevertheless, how to integrate the knowledge from pre-trained models into NMT is another challenge: the improvement in NMT with the standard fine-tuning operation is relatively less~\cite{sun2019baidu}. 
The main reason is that the training objective of the bilingual task is far different from the monolingual task. For example, even with a multi-lingual setting, the objective of BERT or GPT is also predicting the words from the same language, while translation requires the conversion of one language to another. Especially in languages having large differences in morphology and syntax, this gap will lead to that general knowledge from pre-trained models will be erased in the training process of NMT~\cite{yang2019towards}. 
When using the fine-tuning method directly can not work well, \textit{how to explore the potential abilities of pre-trained models in NMT} is an urgent problem to be solved.

In this paper, to address this appealing challenge, we design an \textsc{Apt} framework for acquiring the knowledge from pre-trained models to NMT. Specifically, our \textsc{Apt} framework has two modules. 
First, we propose a \textit{dynamic fusion mechanism} which can learn a task-specific representation by adapting the general representation from pre-trained models, and adopt two controlling methods based on different granularities to fuse the task-specific representation into NMT dynamically. This method could provide rich contextual information for NMT to model sentence better.
Second, we introduce a \textit{knowledge distillation paradigm} to distill the knowledge from pre-trained models to NMT continuously. With this method, NMT could learn the knowledge about how to translate sources sentence to target sentences from parallel data and how to generate a better target sentence from monolingual data in the training process. 
Furthermore, according to our analysis and empirical results, we conclude that the best strategy for using the two methods in the encoder-decoder framework to improve translation quality.

To demonstrate the effectiveness of our \textsc{Apt} approach, we implement the proposed approach based on the advanced pre-trained models and Transformer model~\cite{vaswani2017attention}. It is worth to mention that this framework could be applied to various neural structures based on the encoder-decoder framework~\cite{Bahdanau2015Neural,gehring2016convolutional,vaswani2017attention}.
Experimental results on WMT English to German, German to English and Chinese to English machine translation tasks show that our approach with BERT~\cite{devlin2018bert} or GPT~\cite{radford2018improving} outperforms the Transformer baseline and the fine-tuning counterparts.
    
\section{Background}
\subsection{Neural Machine Translation}
Here, we will introduce neural machine translation based on the Transformer network~\cite{vaswani2017attention}, which has achieved state-of-the-art performance in several language pairs~\cite{deng2018alibaba}.

Denoting a source-target parallel sentence pair as $ \{\textbf{x}, \textbf{y}\} $ from the training set, where $\textbf{x}$ is the source sequence $(x_{1},x_{2},\cdots,x_{i},\cdots,x_I)$ and $\textbf{y}$ is the target sequence $(y_{1},y_{2},\cdots,y_{j},\cdots,y_J)$, $I$ and $J$ are the length of $\textbf{x}$ and $\textbf{y}$, respectively.

In the encoding stage, a multiple layer encoder based on the self-attention architecture is used to encode $ \textbf{x} $ into $\textbf{R}^{E}_{N}$, which is composed by a sequence of vectors $ (\textbf{r}^{E}_{N,1},\textbf{r}^{E}_{N,2},\cdots,\textbf{r}^{E}_{N,i},\cdots, \textbf{r}^{E}_{N,I}) $, $N$ is the depth of the encoder.
The representation $\textbf{R}^{E}_{N}$ is calculated by:
\begin{align}
\label{eq: layernorm}
\textbf{R}^{E}_{N}&=\text{LN}(\textbf{H}^{E}_{N}+\text{FFN}(\textbf{R}^{E}_{N-1})), 
\end{align}
where the $\text{LN}(\cdot)$ and $\text{FFN}(\cdot)$ are layer normalization~\cite{ba2016layer} and feed forward network, respectively. The $\textbf{R}^{E}_{N-1}$ is from the $(N-1)th$ layer. The $\textbf{H}^{E}_{N}$ is computed by:
\begin{align}
\label{eq: attention}
\textbf{H}^{E}_{N}&=\text{Att}(\textbf{Q}^{E}_{N},\textbf{K}^{E}_{N-1},\textbf{V}^{E}_{N-1}),
\end{align}
where the $\text{Att}(\cdot)$ is a self-attention network and the $\textbf{Q}^{E}_{N}$, $\textbf{K}^{E}_{N-1}$, $\textbf{V}^{E}_{N-1}$ are query, key and value matrix, respectively. In this stage, they are equal to $\textbf{R}^{E}_{N-1}$.

Typically, we define that $\textbf{R}^{E}_{0}$ is composed by $\text{emb}(x_{i})$, which is the word embedding of $x_{i}$.

In the decoding stage, the decoder maximizes the conditional probability of generating the $jth$ target word, which is defined as: 
\begin{align}
\label{eq: prob}
    P(y_{j}|y_{<j},\textbf{x})=\text{softmax}(\text{FFN}(\textbf{r}^{D}_{M,j})),
\end{align}
$\textbf{r}^{D}_{M,j}$ is a vector from the target representation matrix $\textbf{R}^{D}_{M}$, $M$ is the depth of the decoder. The $\textbf{R}^{D}_{M}$ is
\begin{align}
\textbf{R}^{D}_{M} = \text{LN}(\text{FFN}(\textbf{S}^{D}_M+\textbf{C}^{D}_M)),
\end{align}
$\textbf{S}^{D}_M$ is computed by Eq. \ref{eq: layernorm}-\ref{eq: attention}, where the query, key and value are equal to $\textbf{R}^{D}_{M-1}$. $\textbf{C}^{D}_M$ is computed by:
\begin{align}
\label{eq: attention2}
\textbf{C}^{D}_{M}&=\text{Att}(\textbf{Q}^{D}_{M},\textbf{K}^{E}_{N},\textbf{V}^{E}_{N}),
\end{align}
where $\textbf{Q}^{D}_{M}$ is equal to $\textbf{S}^{D}_M$, $\textbf{K}^{E}_{N}$ and $\textbf{V}^{E}_{N}$ are equal to $\textbf{R}^{E}_{N}$.

\begin{figure}
    \centering
    \includegraphics[scale = 0.4]{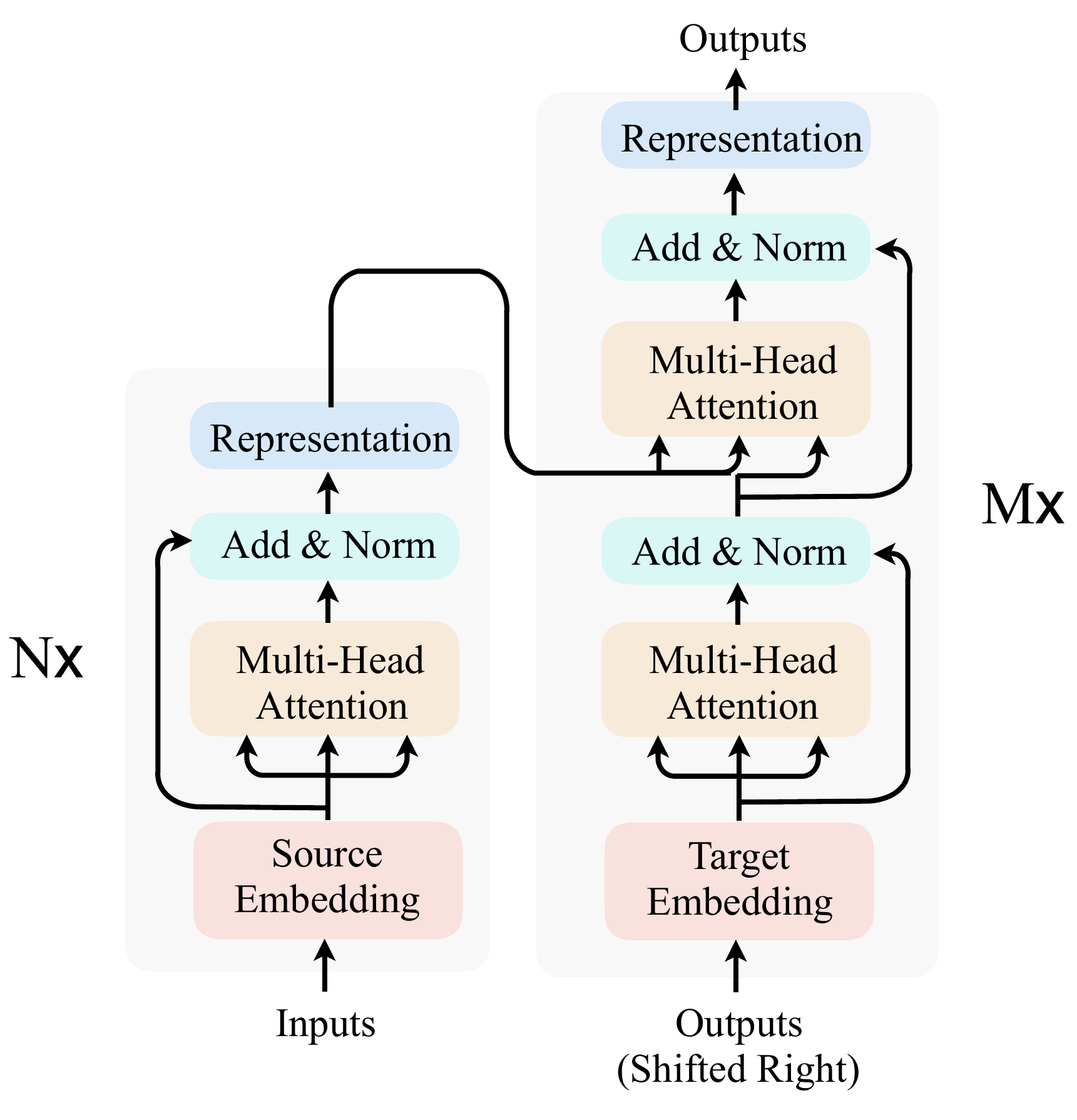}
    \caption{Overview of the structure of the Transformer network~\cite{vaswani2017attention}.}
    \label{fig: transformer}
    \end{figure}

Finally, Transformer is optimized by maximizing the likelihood, denoted by:
\begin{equation}
\label{eq: loss}
\mathcal{L}_{\text{T}}=\frac{1}{J}\sum_{i=1}^{J}\log P(y_{j}|y_{<j},\textbf{x};\theta_{\text{T}}),
\end{equation}
where the $\theta_{\text{T}}$ is the parameters of NMT. 
The overview of the structure of Transformer is shown in Figure \ref{fig: transformer}.

\subsection{Pre-training Model}
Recently, a variety of pre-training models (PT), like ELMo~\cite{peters2018deep}, GPT~\cite{radford2018improving}, BERT~\cite{devlin2018bert}, etc, are proposed to obtain language knowledge from large scale monolingual data. 

Formally, given a sentence $\textbf{z}=(z_{1},z_{2},\cdots,z_{k},\cdots,z_K)$, \textit{K} is the length of \textbf{z}, the pre-trained model is adopted to get the contextual representation: $\textbf{R}^{p}_{L}=\text{PT}(\textbf{z};\theta_{\text{P}})$, where $L$ is the depth of the pre-trained model and $\theta_{\text{P}}$ is the parameters of the pre-training model. The PT($\cdot$) could be implemented by a variety of structures like Bi-LSTM~\cite{peters2018deep} or self-attention network~\cite{radford2018improving,devlin2018bert}.

There are two main objectives to train the pre-trained model~\cite{yang2019xlnet}. The first kind is using an auto-regressive language model objective, which predicts the next word $P(z_{k}|z_{<k};\theta_{\text{P}})$ by the $k$th representation $\textbf{r}^{p}_{L,k}$ from $\textbf{R}^{p}_{L}$. Another popular method is similar to the auto-encoder, which needs to pre-process the sentence \textbf{z} to get a processed one $\hat{\textbf{z}}$ by masking several words $\textbf{z}^{m}$. 
Then, the pre-trained model predicts the masked words to re-construct the $\textbf{z}$ by $P(\textbf{z}^{m}|\hat{\textbf{z}};\theta_{\text{P}})$ in the training process.

\section{Approach}
Owing to the limited amount of parallel data, it is hard for NMT to generate appropriate contextual representation. The pre-trained models are an useful complement to provide NMT models with proper language knowledge.
However, previous integration methods like fine-tuning: initializing parameters from pre-trained models, may not suit for machine translation which is a bilingual generation task. The general contextual information from pre-trained models is quite different from the task-specific representation of NMT model.

Thus, we propose a novel \textsc{APT} framework including a \textit{dynamic fusion mechanism} and a \textit{knowledge distillation paradigm}, to fully utilize pre-trained contextual knowledge in NMT models. We will introduce the two methods in details and discuss the different integration strategies in the encoder and decoder of NMT models. For convenience, we will present the dynamic fusion mechanism on the encoder and the knowledge distillation paradigm on the decoder, respectively.

\subsection{Dynamic Fusion Mechanism} 
We propose a \textit{dynamic fusion mechanism} to obtain the \textit{task-specific representation} by transforming general pre-trained representations in pre-trained models. 
Specifically, we use an adapter for transforming general knowledge to more appropriate features of NMT during the training process.
Furthermore, previous work~\cite{peters2018deep,dou2018exploiting,wang2018multi} shows that representations from each layer in a deep model have different aspect of meaning. Following this intuition, we expand our idea by employing the adapter on all layers' representation from pre-trained models to get different kinds of knowledge, from concrete to abstract.

Formally, the general representations from pre-trained models are 
$ \textbf{R}^{P}=(\textbf{R}^{P}_{1},\cdots,\textbf{R}^{P}_{l},\cdots,\textbf{R}^{P}_{L})
$.
For the $l$th layer's representation $\textbf{R}^{P}_{l}$, the task-specific representation is computed by:
\begin{align}
    \textbf{R}^{T}_{l} &=\text{G}_{l}(\textbf{R}^{P}_{l}), \label{eq: weight}
\end{align}
where the proposed adapter $\text{G}_{l}(\cdot)$ is a simple MLP.
\newcite{mikolov2013exploiting} and \newcite{wu2019learning} pointed out the representation space of similar languages can be transferred by a linear mapping. In our scenario, which is in same language, the mapping function can transfer the general representation to task-specific representation effectively.

\begin{figure}[t]
    \centering
    \includegraphics[scale = 0.60]{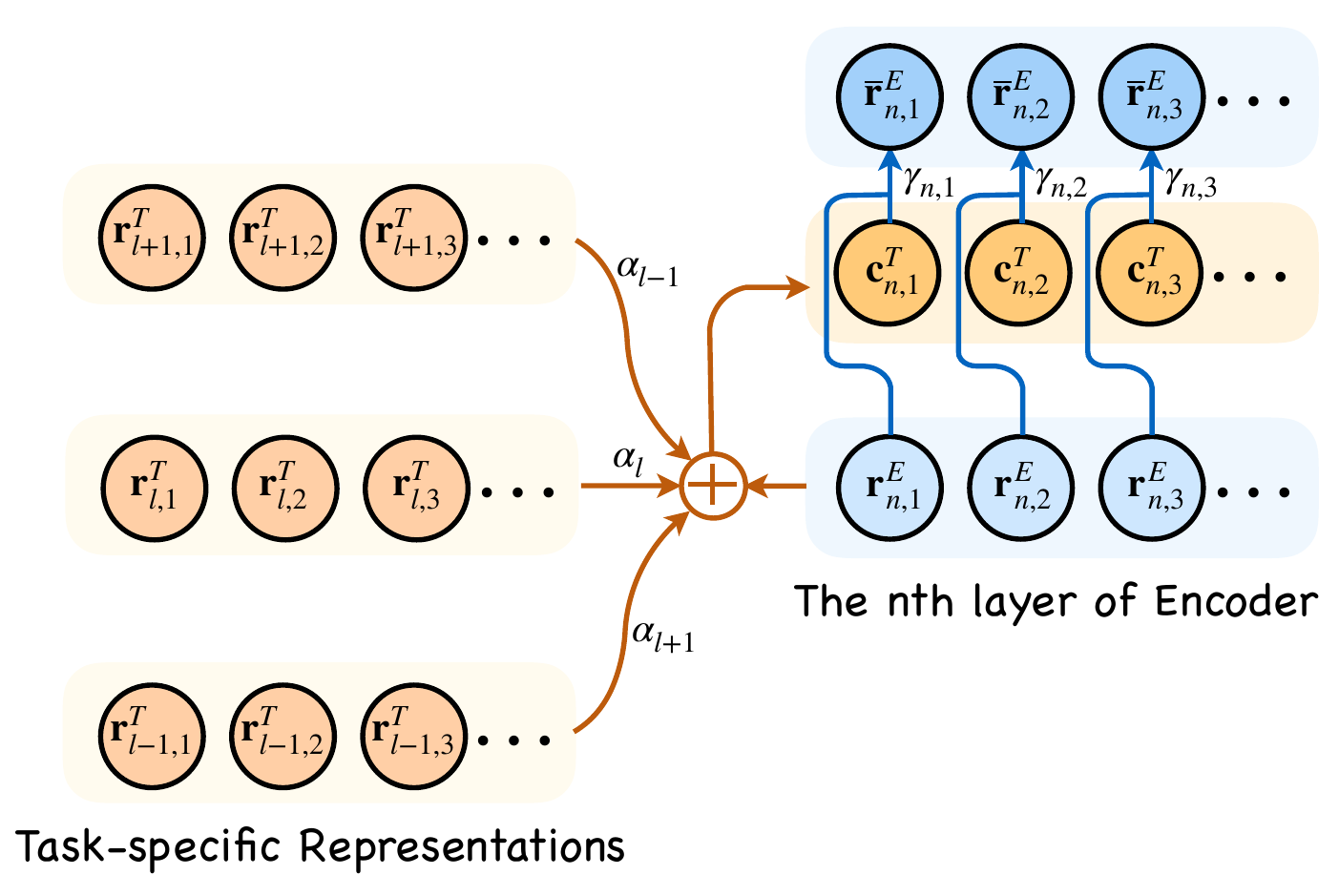}
    \caption{Overview of the dynamic fusion mechanism employed on the encoder of Transformer.}
    \label{fig: src}
    \end{figure}

Subsequently, we propose two methods based on different granularity to control how much the task-specific representation should be fused into Transformer dynamically. 
First, the demand of external information from each layer is different. Thus, compared with using layer coordination~\cite{he2018layer} directly, we further propose a \textit{layer-aware attention mechanism} to capture compound contextual information. 
Formally, given the $n$th layer's vanilla representation $\textbf{R}^{E}_n$ computed by Equation \ref{eq: layernorm}-\ref{eq: attention}, the corresponding external representation is computed by: 

\begin{align}
    \textbf{C}^{T}_n = \sum^{L}_{l=1}\alpha_{l}\textbf{R}^{T}_{l},
    \alpha_{l}=\frac{\text{exp}(e_{l})}{\sum_{t=1}^{L}\text{exp}(e_{t})}, \\
e_{l}=
\text{FFN}(\frac{1}{I}\sum_{i=1}^{I}\textbf{r}^{T}_{l,i}\cdot\frac{1}{I}\sum_{i=1}^{I}\textbf{r}^{E}_{n,i}).
\end{align}
The layer-aware attention mechanism can determine which representation from pre-trained model is more important for current layer. The composite representation $\textbf{C}^{T}_n$ can capture more suitable information by considering a larger context. 

Following above intuition, the demand of each hidden state from same layer is also different. A fine-grained method is necessary to control the fusion ratio of each hidden state. We adopt a simple \textit{contextual gating mechanism}~\cite{kuang2018modeling} to implement it. 

Formally, the representation $\textbf{c}^{T}_{n,i}$ from $\textbf{C}^{T}_{n}$ is fused into the corresponding state $\textbf{r}^{E}_{n,i}$ from $\textbf{R}^{E}_{n}$by:
\begin{align}
\overline{\textbf{r}}^{E}_{n,i} &= \textbf{r}^{E}_{n,i}+\gamma_{n,i} * \textbf{c}^{T}_{n,i}, 
\end{align}
where the gate $\gamma_{n,i}$ is computed by:
\begin{align}
    \gamma_{n,i} &=\text{sigmoid}(\text{FFN}(\textbf{r}^{E}_{n,i}\cdot\textbf{c}^{T}_{n,i}))
\end{align}
The overview is illustrated in Figure \ref{fig: src}.
Different from previous works~\cite{ramachandran2016unsupervised,peters2018deep,radford2018improving}, the proposed feature-based method can make a \textit{deep fusion} which could incorporate appropriate information into each layer, that is, Transformer can access specific surface information in lower layers and the latent one in higher layers. 

\begin{figure}[t]
    \centering
    \includegraphics[scale = 0.6]{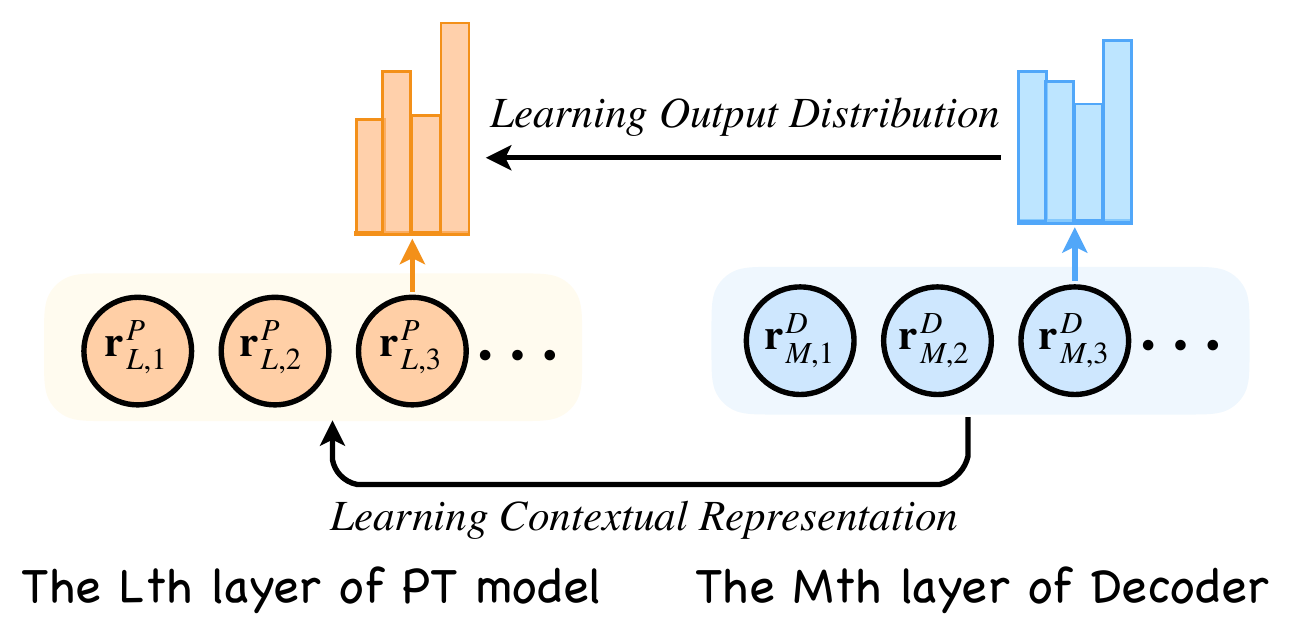}
    \caption{Overview of the knowledge distillation paradigm employed on the decoder of Transformer.}
    \label{fig: trg}
    \end{figure}

\subsection{Knowledge Distillation Paradigm} 
Besides the dynamic fusion mechanism, we also propose a \textit{knowledge distillation paradigm} to learn pre-trained representation in the training process. We introduce two auxiliary learning objectives distilling the knowledge from pre-trained models to NMT in word and sentence levels, respectively.

Firstly, the word level knowledge distillation objective is defined as:
\begin{align}
    \mathcal{L}_{\text{W}}=\frac{1}{J}\sum_{j=1}^{J}\sum_{k=1}^{V}&P(y_{j}=k|\text{y};\theta_{\text{P}}) \nonumber\\ 
    \cdot &\text{log}(P(y_{j}=k|\textbf{x},y_{<j};\theta_{\text{T}}))
\end{align}
where the $J$ is the length of the given target sentence $\textbf{y}$, the $V$ is vocabulary size. 
The $P(y_{j}|y_{<j},\textbf{x};\theta_{\text{T}})$ is computed by Equation \ref{eq: prob}. Compared with only minimizing the one-hot label from reference, this word level training function can learn the output distribution from pre-trained models, which is more diverse.

Then, different from previous sentence level knowledge distillation methods~\cite{chen2017teacher}, our objective learns the sentence level information by fitting contextual representation directly: 
\begin{align}
    \mathcal{L}_{\text{S}}&=\frac{1}{J}||\textbf{R}^{D}_{M}-\textbf{R}^{P}_{L}||^{2}_{2} \nonumber\\&=\frac{1}{J}\sum^{J}_{j=1}||\textbf{r}^{D}_{M,j}-\textbf{r}^{P}_{L,j}||^{2}_{2}, \label{eq: multi-task}
\end{align}
where the $M$ is output layer of the decoder. The $\textbf{r}^{D}_{M,j}$ and $\textbf{r}^{P}_{L,j}$ are from the decoder and pre-trained model, respectively. The vanilla sentence level training objective need to sample output sentence in the training process, which may cause a bias and decrease efficiency, while our method could learn contextual information from the hidden state directly. The overview of the proposed knowledge distillation paradigm is shown in Figure \ref{fig: trg}.

Finally, the loss function of our \textsc{Apt} is:
\begin{align}
\mathcal{L}=\mathcal{L}_{\text{T}}+\eta \cdot \mathcal{L}_{\text{S}}+\beta \cdot \mathcal{L}_{\text{W}}, \label{eq: joint}
\end{align}
where $\eta$ and $\beta$ are used to balance the preference among the two losses, which we are set to 0.5 individually.

\subsection{Integration Strategy}
The duties of encoder and decoder in NMT are different.  
And the decoder has two states in the training and inference stages. Thus, different integration strategies are needed for our proposed approach.

The encoder needs to capture contextual information by modeling input sentence. The goal of exploiting external contextual information in the encoder is for modeling input sentence better. Thus, even the knowledge distillation method with sentence level objective could be used in the encoder, the dynamic fusion mechanism is more suitable.

However, compared with the encoder, the decoder is difficult to exploit pre-trained knowledge for two reasons. First, the main role of the decoder is generating a target sentence by feeding source representation, which involves the transformation of semantic space. So, the representation from the decoder is far different from the pre-trained model. 
Then, the \textit{exposure bias}~\cite{lee2018deterministic,wu2018beyond} leads to that the ground-truth representation which is generated by reference is not available in the inference stage. 
So, we think using the knowledge distillation to learn language knowledge is a better solution, which will not influence the original goal of translation, and help to generate a better sentence.

In general, we integrate the \textsc{Apt} framework by employing the dynamic fusion mechanism on the encoder and the knowledge distillation paradigm on the decoder. We also report the comparison of other strategies in the experiment.

\section{Experiment}

\begin{table*}[t]
    \centering
    \begin{tabular}{l|cc||cc|cc|cc}
    \hline
    \multirow{2}{*}{Model}&\multicolumn{2}{c||}{Pre-trained Model} &\multicolumn{2}{c|}{EN$\rightarrow$DE}&\multicolumn{2}{c|}{DE$\rightarrow$EN} & \multicolumn{2}{c}{ZH$\rightarrow$EN}  \\
    
    
    &Encoder&Decoder&BLEU&$\Delta$&BLEU&$\Delta$&BLEU&$\Delta$  \\
    \hline
    Transformer~\cite{vaswani2017attention}&N/A&N/A&27.3	&$-$&N/A &$-$& N/A&$-$\\
    Transformer~\cite{zheng2019dynamic}&N/A&N/A&27.14	&$-$&N/A &$-$& N/A&$-$\\
    Transformer~\cite{dou2018exploiting}&N/A&N/A&27.31	&$-$&N/A &$-$& 24.13&$-$\\
    \hline
    Transformer&N/A&N/A&27.31	&$-$&32.51 &$-$& 24.47&$-$\\
    \hline
    
    
    \ \ \multirow{10}{*}{\textit{w/} Fine-tuning}&GPT & N/A&27.82 &+0.51 &33.17&+0.66 &25.11 &+0.64 \\
    &N/A &GPT&27.45 &+0.14 &32.87&+0.36 &24.59 &+0.12\\
    &GPT&GPT&27.85 &+0.54 &32.79&+0.28 &25.21 &+0.74 \\
    \cline{2-9}
    &BERT&N/A &28.22 &+0.91 &33.64&+1.13 &25.33 &+0.86 \\
    &N/A &BERT&27.42 &+0.11 &33.13&+0.62 &24.78 &+0.31 \\
    &BERT &BERT&28.32 &+1.01 &33.57&+1.06 &25.45 &+0.98 \\
    \cline{2-9}
    &GPT&BERT&28.29&+0.98&33.33&+0.82&25.42&+0.95 \\
    &BERT&GPT&28.32 &+1.01 &33.57&+1.05 &25.46 &+0.99 \\
    \cline{2-9}
    &\multicolumn{2}{c||}{MASS} &28.07 &+0.76 &33.29&+0.78 &25.11 &+0.64\\
    &\multicolumn{2}{c||}{DAE} &27.63 &+0.33 &33.03&+0.52 &24.67 &+0.20\\
    \hline
    \ \ \multirow{4}{*}{\textit{w/} \textsc{Apt} Framework}&GPT&BERT &28.89 &+1.58 &34.32&+1.81 &25.98 &+1.51 \\
    &BERT&GPT &\textbf{29.23} &\textbf{+1.92} &\textbf{34.84}&\textbf{+2.33}&26.21 &+1.74 \\
    &GPT&GPT &28.97 &+1.66 &34.26&+1.75 &26.01 &+1.54\\
    &BERT&BERT &29.02 &+1.71 &34.67&+2.16 &\textbf{26.46}&\textbf{+1.99} \\
    \hline
    \end{tabular}
    \caption{Translation qualities on the EN$\rightarrow$DE, DE$\rightarrow$EN and ZH$\rightarrow$EN experiments.}
    \label{tab: mt}
    \end{table*}

\subsection{Implementation Detail}
\paragraph{Data-sets} We conduct experiments on the WMT data-sets\footnote{http://www.statmt.org/wmt17/translation-task.html}, including WMT17 Chinese to English (ZH$\rightarrow$EN), WMT 14 English to German (EN$\rightarrow$DE) and German to English (DE$\rightarrow$EN) and the corresponding monolingual data.

On the ZH$\rightarrow$EN, we use WMT17 as training set which consists of about 7.5 million sentence pairs (only CWMT part). We use {\texttt{newsdev2017}} as validation set which has 2002 sentence pairs, and \texttt{newstest2017} as test set which have 2001 sentence pairs.
On the EN$\rightarrow$DE and DE$\rightarrow$EN, we use WMT14 as training set which consists of about 4.5 million sentence pairs. We use {\texttt{newstest2013}} as validation set which has 3000 sentence pairs, and \texttt{newstest2014} as test set which have 3003 sentence pairs.

Following \newcite{song2019mass}, on the English and German, we use the monolingual data from WMT News Crawl. We select 50M sentence from year 2007 to 2017 for English and German respectively. Then, we choose 50M sentence from Common Crawl for Chinese.

\paragraph{Settings} We apply byte pair encoding (BPE)~\cite{sennrich2015neural} to all language pairs and limit the vocabulary size to 32K. 

For Transformer, we set the dimension of the input and output of all layers as 512, and that of the feed-forward layer to 2048. 
We employ 8 parallel attention heads. The number of layers for the encoder and decoder are 6. 
Sentence pairs are batched together by approximate sentence length. Each batch has 50 sentence and the maximum length of a sentence is limited to 100. 
We use label smoothing with value 0.1 and dropout with a rate of 0.1. We use the Adam~\cite{kingma2014adam} to update the parameters, and the learning rate was varied under a warm-up strategy with 4000 steps. Other settings of Transformer follow \newcite{vaswani2017attention}. 

we also implement GPT~\cite{radford2018improving}, BERT~\cite{devlin2018bert} and MASS~\cite{song2019mass} in our Transformer system. The implementation details are as follows:
\begin{itemize}
    \item GPT: \newcite{radford2018improving} proposed a pre-trained self-attention language model. We implement it on both source and target languages based on the aforementioned Transformer decoder.
    \item BERT: \newcite{devlin2018bert} proposed a pre-trained bi-directional encoder optimized by the masked token and next sentence objectives. Following \newcite{lample2019cross}, we implement it only using the masked token objective, which doesn't require monolingual data has document boundary.
    \item MASS: \newcite{song2019mass} proposed a masked sequence to sequence pre-training model for text generation tasks. It masks a continuous segment from a sentence as the label, and the rest of the sentence as the input of encoder. We implement it in our Transformer system without any modification.
\end{itemize}

After the training stage, we use beam search for heuristic decoding, and the beam size is set to 4. 
We measure the translation quality with the NIST-BLEU~\cite{Papineni2002bleu}. 
We implement our approach with the in-house implementation of Transformer derived from the \textit{tensor2tensor}\footnote{https://github.com/tensorflow/tensor2tensor}.

\subsection{Main Results}
\paragraph{Translation Quality} The results on the EN$\rightarrow$DE, DE$\rightarrow$EN and ZH$\rightarrow$EN are shown in Table \ref{tab: mt}. For a fair comparison, we also report several Transformer baseline from previous work~\cite{vaswani2017attention,zheng2019dynamic,dou2018exploiting}. Our Transformer baseline achieves similar or better results comparing with them. Compared with our baseline, Transformer with the \textsc{Apt} framework based on different pre-trained models improves 1.92, 2.33 and 1.99 BLEU scores on the EN$\rightarrow$DE, DE$\rightarrow$EN and ZH$\rightarrow$EN, respectively (\textbf{bold} font). It's worth to mention that the percentage improvement on the ZH$\rightarrow$EN, whose difference of syntax and morphology is bigger than German and English, is more than other language pairs.  

\paragraph{Compared with Fine-tuning} We also implement the fine-tuning method with different pre-trained models. When the encoder is initialized by BERT and the decoder is initialized by BERT or GPT, the BLEU score improves about 1 point on three translation tasks. Our \textsc{Apt} framework outperforms the fine-tuning method on all tasks whenever using BERT or GPT. This results demonstrate that the proposed approach is more effective for obtaining the knowledge from pre-trained model than fine-tuning in neural machine translation. 

\paragraph{GPT Vs. BERT}
Although our work combining with GPT or BERT achieves remarkable improvements, there are several differences when employing them on encoder or decoder. First, BERT is better than GPT on the encoder when using the proposed \textsc{Apt} framework (+0.13 to +0.48). We think the reason is that compared with the uni-directional language model of GPT, the masked language model could obtain more contextual information. While on the decoder side, GPT gets better performance than BERT due to it can model sequential information which is an important factor for the decoding process.

\begin{table}[t]
    \centering
    \begin{tabular}{l||c}
    \hline
    \textbf{Model} & BLEU  \\
    
    \hline
    Transformer-Big~\cite{vaswani2017attention} &28.46 \\
    \ \ w/ Fine-tuning~\cite{lample2019cross}& 27.70 \\
    \ \ w/ Feature~\cite{lample2019cross} & 28.70 \\
    \ \ w/ \textsc{CTnmt}~\cite{yang2019towards} & 30.10\\
    \hline
    Transformer-Base &27.31 \\
    \ \ w/ \textsc{Apt} framework & 29.23 \\
    \hline
    \end{tabular}
    \caption{The comparison of the proposed method and previous work on the EN$\rightarrow$DE task.}
    \label{tab: compare}
    \end{table}

\begin{table}[t]
    \centering
    
    \begin{tabular}{l||c|c}
    \hline
    Model  &BLEU&$\Delta$ \\
    \hline
    Transformer  &27.31&$-$ \\
    \hline
    \ \ \textit{w/o} Knowledge Distillation  &28.77 &+1.46 \\
    \ \ \ \ \textit{w/o} Contextual Gating  &28.44 &+1.13\\
    \ \ \ \ \textit{w/o} Layer-aware Attention   &28.39 &+1.08 \\
    \tabincell{l}{\ \ \ \ \textit{w/o} Contextual Gating \\ 
    \ \ \ \ \ \ \textit{w/o} Layer-aware Attention} &28.03&+0.72  \\
    \hline
    \ \ \textit{w/} Knowledge Distillation   &29.23 &+1.92 \\
    \ \ \ \ \textit{w/o} Contextual Gating  &28.91 &+1.60 \\
    \ \ \ \ \textit{w/o} Layer-aware Attention   &28.68 &+1.37 \\
    \tabincell{l}{\ \ \ \ \textit{w/o} Contextual Gating \\ 
    \ \ \ \ \ \ \textit{w/o} Layer-aware Attention} &28.43&+1.12  \\
    \hline
    \hline
    \ \ \textit{w/o} Dynamic Fusion  &28.68 &+1.37 \\
    \ \ \ \ \textit{w/o} Word Distillation   &28.31 &+1.02 \\
    \ \ \ \ \textit{w/o} Sent Distillation   &28.56 &+1.15 \\
    \hline
    \ \ \textit{w} Dynamic Fusion  &29.23 &+1.92 \\
    \ \ \ \ \textit{w/o} Word Distillation   &28.76 &+1.45 \\
    \ \ \ \ \textit{w/o} Sent Distillation   &28.87 &+1.56 \\
    \hline
    \end{tabular}
    \caption{Ablation study on the EN$\rightarrow$DE task.}
    \label{tab: ablation}
    \end{table}

\paragraph{Compared with Previous Work}
We also report several recent work related to use pre-trained model in NMT. 
The results are summarized in Table \ref{tab: compare}. When using the public BERT\footnote{https://github.com/google-research/bert} to fine-tune the Transformer-big~\cite{vaswani2017attention}, BLEU score decreases 0.76. However, in our implementation, the fine-tuning method improves 1.01 BLEU based on the Transformer-base. Furthermore, the feature-based approach is better than fine-tuning which contrasts other tasks~\cite{devlin2018bert}. This result also verified the fine-tuning doesn't fit NMT.
Our approach with \textit{base} setting could outperform their whose parameter size is far larger than us\footnote{\textsc{CTnmt} uses the \textit{big} setting and beam size is 8. Furthermore, the size of monolingual data they used is far larger than us. So, it's unfair to compare us with them directly.}. 

\subsection{Ablation Study}
To show the effectiveness of each module from the proposed framework, we do a detailed ablation study here. On the one hand, we show the effectiveness of the context gating and layer-aware attention from the dynamic fusion in the first two parts. 
Whether using knowledge distillation or not, the layer-aware attention is a bit more important than context gating. The fine-grained method of context gating could provide further improvement based on the layer-aware attention. When ablating both of them, which like a layer coordination method, the BLEU score drops about 0.7 point.

On the other hand, the word level and sentence level distillation objectives could be used individually. We also evaluate them with or without dynamic fusion mechanism.
Compared with word level distillation, without sentence level distillation will have more negative influence, which reveal that learning contextual knowledge is important than only learning the output distribution.

\begin{table}[t]
    \centering
    \begin{tabular}{l||c|c}
    \hline
    \textbf{Model} &\textbf{Method}&\textbf{BLEU} \\
    \hline
    Transformer &N/A&27.31 \\
    \hline
     \ \ \multirow{3}{*}{Encoder}&w/ Dynamic Fusion &28.77  \\
     \cline{2-3}
     & w/ Knowledge Distillation &28.21 \\
     \cline{2-3}
     & w/\tabincell{c}{ Dynamic Fusion \\ Knowledge Distillation} &28.69\\
    \hline
    \hline
    \ \ \multirow{3}{*}{Decoder} &w/ Dynamic Fusion &  27.41\\
    \cline{2-3}
    & w/ Knowledge Distillation & 28.68  \\
    \cline{2-3}
    & w/\tabincell{c}{ Dynamic Fusion \\ Knowledge Distillation} &  27.71 \\
    \hline
    \end{tabular}
    \caption{The comparison of translation qualities for using the dynamic fusion mechanism and knowledge distillation paradigm with different strategies on the EN$\rightarrow$DE task.}
    \label{tab: strategy}
\end{table}

\subsection{Impact of Different Integration Strategies}
In this section, we analyze the different integration strategies for our \textsc{Apt} framework. Specifically, we employ three integration settings on the encoder and decoder, respectively. The results are summarized in Table \ref{tab: strategy}. 

\paragraph{Encoder side} The different strategies employed on the encoder are shown in the first part of Table \ref{tab: strategy}. Here, we only use sentence level knowledge distillation. The knowledge distillation can help the encoder to model input sentence better, while the effect is not as good as using the dynamic fusion mechanism.
Moreover, the performance doesn't improve when adopting both the knowledge distillation and the dynamic fusion. It shows that the effectiveness of knowledge distillation is covered by dynamic fusion in this scenario.
\paragraph{Decoder side} 
We make a comparison on the decoder side under the same settings above mentioned. 
The dynamic fusion doesn't work on the decoder side, in which pre-trained models can't get the ground truth as a input in the decoding stage, so the task-specific representation generated by the dynamic fusion is incomplete and contains noisy.
According to this experiment, the knowledge distillation is better than feature-based or fine-tuning methods in the decoder. 

\subsection{Effectiveness on Different Layers}
We compare the effectiveness of employing the \textsc{Apt} on different layers. The results are shown in Table \ref{tab: src_layer}. The phenomena from the encoder and decoder sides are different. On the encoder side, more layers fuse external knowledge, better performance the model achieve. Moreover, high layers can get more gain comparing with low layers. These results indicate that the dynamic fusion can improve the ability of modeling input sentence at all layers. High layers of the encoder need external contextual knowledge more than low layers to get the semantic from the input sentence.

On the decoder side, adopting knowledge distillation on the output layer obtains the best performance. And only adopting it on the embedding is better than others. We think the middle layers focus on transforming the source representation from the encoder to the target, so the external contextual knowledge doesn't help them much. On the embedding layer, the pre-trained embedding is better than the embedding of NMT. So, only fitting embedding can get a considerable improvement.  The representation from output layer is used to generate the target sentence, the contextual representation learned from the  pre-trained model could give it more language information to generate a better sentence.

\begin{table}[t]
    \centering
    \begin{tabular}{l||c|c}
    \hline
    \textbf{Model} &Layers & BLEU  \\
    \hline
    Transformer &N/A&27.31 \\
    \hline
    \multirow{4}{*}{\ \ \textit{w/} Dynamic Fusion} 
     &Embedding&27.94 \\
     \cline{2-3}
     &1st-5th&28.42 \\
     \cline{2-3}
     &Output&28.34 \\
     \cline{2-3}
     &All&28.77 \\
     \hline
     \hline
     \multirow{4}{*}{\ \ \textit{w/} Knowledge Distillation} 
      &Embedding&28.21 \\
      \cline{2-3}
      &1st-5th&28.01 \\
      \cline{2-3}
      &Output&28.68 \\
      \cline{2-3}
      &All&28.22 \\
    \hline
    \end{tabular}
    \caption{The comparison of employing dynamic fusion mechanism and knowledge distillation paradigm on different layers on the EN$\rightarrow$DE task.}
    \label{tab: src_layer}
\end{table}

\section{Related Work}
\paragraph{Pre-trained Model}
In NLP field, there are many pre-trained models have been proposed to learn the contextual information from large scale monolingual data.
\newcite{peters2018deep} introduced Embedding learned from Bi-LSTM based Language Models (ELMo) and successfully apply it in question answering, textual
entailment, etc. 
Inspired by them, \newcite{radford2018improving} proposed to use self-attention network based language model (GPT) to replace the Bi-LSTM structure, which further improves the performance of the pre-trained model.
Then, \newcite{devlin2018bert} proposed to use the bi-directional encoder representation from Transformer (BERT) with a masked token objective and a next sentence objective to capture global contextual information. After that, several varieties are proposed, like MASS~\cite{song2019mass}, XLNet~\cite{yang2019xlnet}, etc.

These pre-training methods attended the state-of-the-art in several tasks by fine-tuning the pre-trained parameters with labeled data from downstream tasks.
However, fine-tuning these advanced model by parallel data can not work well in NMT, because of the gap between the bilingual and monolingual tasks. Our \textsc{APt} framework can erase this gap and fully release the potential of these model in NMT filed.

\paragraph{Exploiting Monolingual Data for NMT}
Several successful attempts have been made to utilize monolingual data in NMT directly.
\newcite{sennrich2016improving} proposed to use back-translation to generate synthetic parallel data from monolingual data.
\newcite{currey2017copied} proposed a copy mechanism to copy fragments of sentences from monolingual data to translated outputs directly.
\newcite{zhang2018joint} proposed to jointly train the source-to-target and target-to-source NMT models with the pseudo parallel data from monolingual data.
However, these studies only focus on the usage of word or surface information, the rich contextual information from large scale monolingual data does not be fully utilized.

Then, some researchers also pay attention to apply the pre-trained model for NMT. \newcite{di2017can} use source side pre-trained embedding and integrate it into NMT with a mix-sum/gating mechanism. They only focus on utilizing the pre-trained embedding, leaving the underlying linguistic information ignored.
\newcite{ramachandran2016unsupervised} firstly proposed the fine-tuning method in NMT. However, the general information can not be used in NMT directly leading to the information from pre-trained models is less exploited.
Our approach can fuse the adapted task-specific representation into NMT and distill knowledge from pre-trained model to NMT in the training process, exploiting the pre-trained knowledge effectively.
\section{Conclusion}
In this paper, we first address the problem that the contextual knowledge from pre-trained models can not be used well with the fine-tuning method in NMT, due to the large gap between the bilingual machine translation task and monolingual pre-trained models.
Then, we propose an \textsc{Apt} framework, which could fuse the task-specific representation adapted from general representation by a dynamic fusion mechanism and learn the contextual knowledge from pre-trained models by a knowledge distillation paradigm. Moreover, we conclude the best strategy for using the two methods in the encoder-decoder framework.
Experiments on three machine translation tasks show that the proposed \textsc{Apt} framework achieves prominent improvements by fully acquiring the knowledge from pre-trained models to NMT. 
\section*{Acknowledgements}
We would like to thank the anonymous reviewers for their insightful comments. Thanks to Shaohui Kuang and Wei Zou for their insightful comments. Shujian Huang is the corresponding author.
This work is supported by the National Science Foundation of China (No. U1836221, 61672277)and the Jiangsu Provincial Research Foundation for Basic Research (No. BK20170074).
\bibliography{aaai2020}
\bibliographystyle{aaai}
\end{document}